\documentclass[letterpaper]{article} 
\usepackage{aaai25}  
\usepackage{times}  
\usepackage{helvet}  
\usepackage{courier}  
\usepackage[hyphens]{url}  
\usepackage{graphicx} 
\usepackage{natbib}  
\usepackage{caption} 
\usepackage{algorithm}
\usepackage{algorithmic}

\usepackage{newfloat}
\usepackage{listings}
\DeclareCaptionStyle{ruled}{labelfont=normalfont,labelsep=colon,strut=off} 
\floatstyle{ruled}
\newfloat{listing}{tb}{lst}{}
\floatname{listing}{Listing}
\nocopyright 

\usepackage{times}
\usepackage{latexsym}

\usepackage{microtype}

\usepackage{amssymb}
\usepackage{bm}
\usepackage{times}

\usepackage{graphicx}
\usepackage{multirow}
\usepackage{enumerate}
\usepackage{booktabs}
\usepackage{enumitem}
\usepackage{amsmath}
\usepackage{xcolor}
\usepackage{bbding}

\title{Affordances-Oriented Planning using Foundation Models for Continuous Vision-Language Navigation}

\author{Jiaqi Chen$^{1}$ ~ Bingqian Lin$^{2}$ ~ Xinmin Liu$^{3}$ ~ Lin Ma$^{3}$ ~ Xiaodan Liang$^{2\ast}$ ~ Kwan-Yee K. Wong$^{1}$\thanks{Corresponding authors.}  \\
{$^1$The University of Hong Kong} ~~ $^2$Shenzhen Campus of Sun Yat-sen University ~~ $^3$Meituan \\
}

\usepackage{bibentry}

\begin{document}

\maketitle

\begin{abstract}

LLM-based agents have demonstrated impressive zero-shot performance in vision-language navigation (VLN) task. However, existing LLM-based methods often focus only on solving high-level task planning by selecting nodes in predefined navigation graphs for movements, overlooking low-level control in navigation scenarios. To bridge this gap, we propose \textbf{AO-Planner}, a novel \textbf{A}ffordances-\textbf{O}riented \textbf{Planner} for continuous VLN task. Our AO-Planner integrates various foundation models to achieve affordances-oriented low-level motion planning and high-level decision-making, both performed in a zero-shot setting. Specifically, we employ a \textbf{V}isual \textbf{A}ffordances \textbf{P}rompting (\textbf{VAP}) approach, where the visible ground is segmented by SAM to provide navigational affordances, based on which the LLM selects potential candidate waypoints and plans low-level paths towards selected waypoints. 
We further propose a high-level PathAgent which marks planned paths into the image input and reasons the most probable path by comprehending all environmental information.
Finally, we convert the selected path into 3D coordinates using camera intrinsic parameters and depth information, avoiding challenging 3D predictions for LLMs.
Experiments on the challenging R2R-CE and RxR-CE datasets show that AO-Planner achieves state-of-the-art zero-shot performance (8.8\% improvement on SPL). 
Our method can also serve as a data annotator to obtain pseudo-labels, distilling its waypoint prediction ability into a learning-based predictor. This new predictor does not require any waypoint data from the simulator and achieves 47\% SR competing with supervised methods.
We establish an effective connection between LLM and 3D world, presenting novel prospects for employing foundation models in low-level motion control.

\end{abstract}

\section{Introduction}
\label{sec:intro}

\begin{figure}[t]
\begin{center}
 \includegraphics[width=0.96\linewidth]{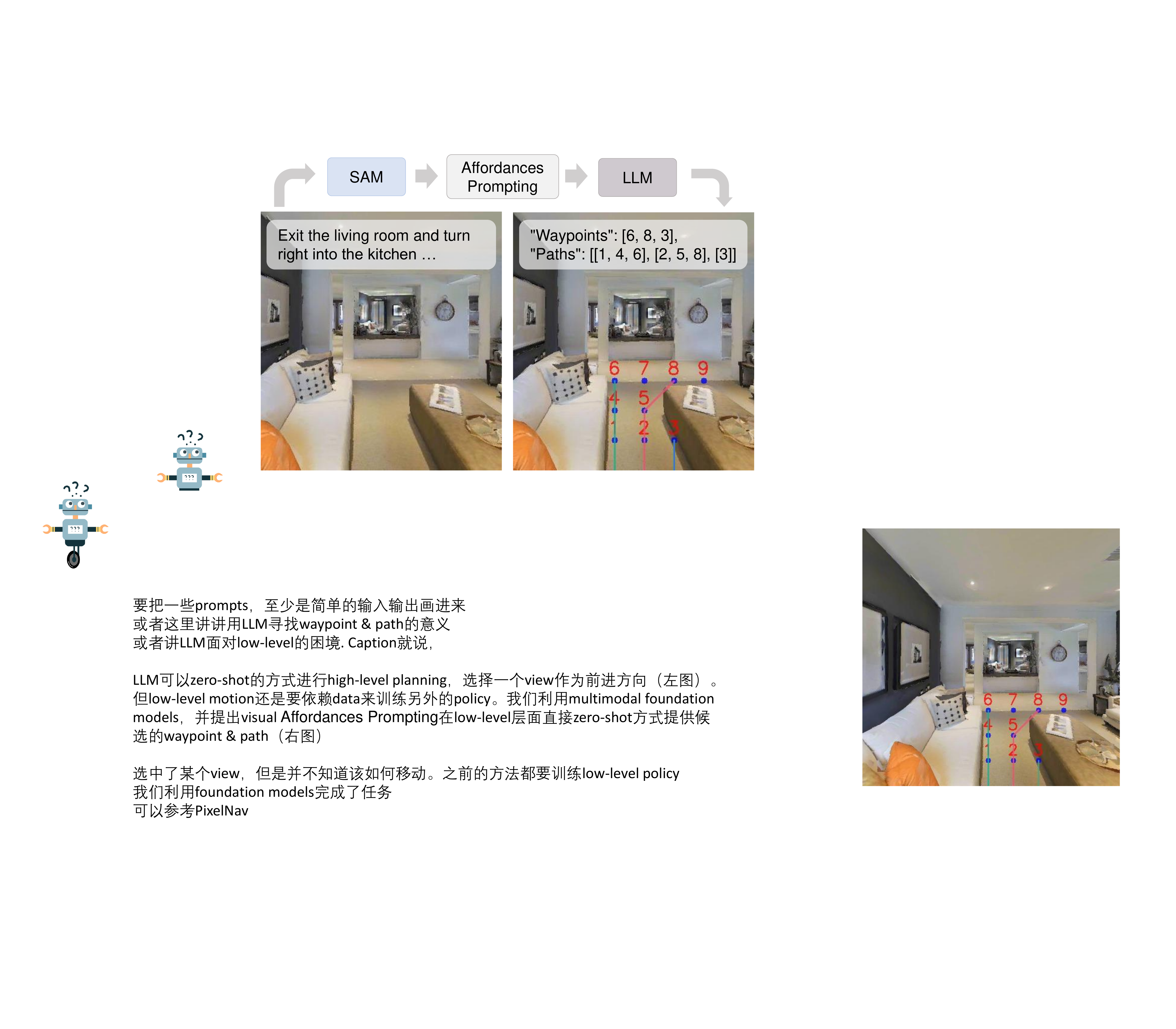}
\end{center}
\vspace{-5mm}
  \caption{
 In discrete VLN, LLMs only need to perform high-level planning by selecting a view as the forward direction (left). For continuous environments, previous agents rely on collecting simulator data to train low-level policies. In this paper, we utilize multimodal foundation models and propose visual affordances prompting to predict low-level candidate waypoints and paths in a zero-shot setting (right).
  }
\vspace{-2mm}
\label{fig:intro}
\end{figure}

Large Language models (LLMs) \cite{openai2023gpt4,openai2023gpt4v-system,team2023gemini,touvron2023llama2,anil2023palm} have shown strong capabilities and potentials in embodied intelligence tasks.
In particular, for vision-and-language navigation (VLN) task \cite{anderson2018vision,qi2020Object,hong2021vln,chen2022think,qiao2022hop,wang2023scaling}, where agents need to make decisions by comprehending language instructions and visual observations, LLMs have demonstrated impressive zero-shot navigation performance. Recent works \cite{zhou2023navgpt,long2023discuss,chen2024mapgpt} have effectively leveraged the high-level capabilities of LLMs for task planning, decision-making, instruction parsing, etc. 

However, there exist several challenges and gaps which hinder the application of LLM-based agents to more practical navigation tasks.
For instance, agents in the classical discrete VLN setting only need to navigate between predefined viewpoints with high-level decision-making capabilities. 
In contrast, for more practical VLN in continuous environments (VLN-CE), learning-based models \cite{krantz2020beyond,krantz2022sim,hong2022bridging,wang2023dreamwalker,an2024etpnav,wang2024lookahead} must predict low-level actions (e.g., turn left $15^\circ$) or positions of subgoals in the environment that can drive agents in the simulator.
Recently proposed zero-shot agents for navigation in continuous environments \cite{cai2023bridging,chen20232} only utilize LLMs for high-level task planning, and they still require a large amount of sampled environmental data from the simulator to train low-level policies. 
Obviously, the gaps between discrete and continuous environments have made current research inadequate in exploring and exploiting the practical navigation potentials of LLMs. Hence, we raise the following question:
\textit{can LLMs not only handle high-level tasks but also serve as low-level motion planners?}

In this paper, we propose a novel zero-shot affordances-oriented planning framework, named \textbf{AO-Planner}, which can navigate agents in continuous environments and is capable of handling both low-level path planning and high-level decision-making. Our key target is to bridge the gap between LLM's RGB space predictions and 3D world navigation via pixel-level path planning in \textit{affordances}.
First, we utilize Grounded SAM \cite{ren2024grounded} to segment navigable regions on the ground as navigational affordances and sample points for the LLM to choose from. 
We then propose a novel Visual Affordances Prompting (\textbf{VAP}) method to leverage LLM's spatial understanding abilities and achieve low-level motion in a zero-shot manner. 
As shown in Figure~\ref{fig:intro},
the LLM \cite{team2023gemini,reid2024gemini} is tasked with selecting the most appropriate waypoints from affordances and connecting some of the points in affordances to plan navigable paths while avoiding obstacles. This process is conducted entirely in the RGB space, in which LLM excels, without the necessity to predict 3D spatial coordinates directly. These low-level candidates (including waypoints and the corresponding paths) are then visualized and fed to another zero-shot PathAgent in the second stage, which is responsible for selecting the path to follow. Finally, by combining depth information with the camera intrinsic parameters, pixel predictions in the RGB space are converted into a series of 3D coordinates that navigate the agent to the designated location.

Experiments on the R2R-CE \cite{krantz2020beyond} and RxR-CE \cite{ku2020room} datasets demonstrate that our proposed framework achieves state-of-the-art zero-shot performance. The previous best-performing method, $A^2$Nav \cite{chen20232}, still relies on sampling a large amount of data from the simulator to train low-level policies. In contrast, our approach has shown an improvement of 2.9\% in SR on R2R-CE and  5.6\% in SR on RxR-CE, without using any training data. For the SPL metric on RxR-CE, our method can even achieve an 8.8\% performance improvement.
To simulate the challenges of motion planning in the continuous Habitat simulator \cite{savva2019habitat}, we also adopt a setting that avoids using any additional data, such as the navigation graph from discrete MP3D simulator \cite{chang2017matterport3d}, to train low-level policies.
We implement a variant of the AO-Planner where we utilize waypoints predicted by LLMs as pseudo-labels, replacing the navigation graph obtained from the MP3D simulator as ground truth, to train a new learning-based waypoint predictor at the same data scale. We combine this distilled waypoint predictor with the VLN agent used in ETPNav \cite{an2024etpnav} and achieve competitive performance, with an SR of 47\%.
The tremendous potentials of multimodal LLM in low-level motion planning are revealed through AO-Planner's impressive performance.

Our main contributions can be summarized as follows.
\begin{itemize}
\setlength{\itemsep}{0pt}
\setlength{\parsep}{0pt}
\setlength{\parskip}{0pt}
  \item 
  We propose AO-Planner for the VLN-CE task. It leverages foundation models for affordances-oriented planning and converts predictions in the RGB space to 3D coordinates, 
  bridging the gap between LLM's high-level decision-making and 3D world navigation.
  \item 
  We present a novel visual affordances prompting (VAP) method to unleash spatial understanding and reasoning abilities of LLMs. This also uncovers previously unexplored motion planning abilities of LLMs.
  \item 
   Our AO-Planner achieves state-of-the-art zero-shot performance without using any simulator data for training low-level policies. It can also provide reliable pseudo-labels for training supervised models to achieve competitive performance.

\end{itemize}
\section{Related Work}

\paragraph{Vision-and-Language Navigation (VLN)}
VLN \cite{anderson2018vision} is a representative task in the field of embodied AI, requiring an agent to combine instructions and visual observations to move through complex environments and reach a target location. 
In this task, 
the most classic scenario is the discrete VLN setting based on the MP3D simulator \cite{chang2017matterport3d}, where an agent only needs to select a node from a predefined navigation graph and teleport to it. Therefore, previous works \cite{wang2019reinforced,ma2019self, deng2020evolving,tan2019learning,hong2021vln,chen2021history,chen2022think,qiao2022hop,guhur2021airbert,an2022bevbert,qiao2023march,wang2023scaling} have focused on endowing agents with high-level decision-making abilities without considering low-level motion.

In recent years, VLN in continuous environment (VLN-CE) built on the Habitat simulator \cite{savva2019habitat} has received much attention. In this setting, the action space only consists of low-level actions, such as forward and rotate, instead of teleporting the agent directly. To overcome the challenge of directly predicting low-level actions, some works \cite{krantz2021waypoint,irshad2021hierarchical,hong2022bridging,krantz2022sim,an2024etpnav} train a waypoint model to predict candidate waypoints in the surroundings as a replacement for the candidate waypoints provided by the MP3D simulator, bridging the gap between discrete and continuous VLN. However, such waypoint models rely on the simulator-specific navigation graph data from MP3D for training, and may suffer from generalization issues. Furthermore, the candidate positions predicted by these models are only some locations near the agent, which can be reached simply by moving forward without considering motion planning. 
In this paper, we propose a zero-shot affordances-oriented planning framework (AO-Planner) driven by foundation models for low-level motion planning, which does not rely on any simulator data for training.

\begin{figure*}[t]
\begin{center}
 \includegraphics[width=0.98\linewidth]{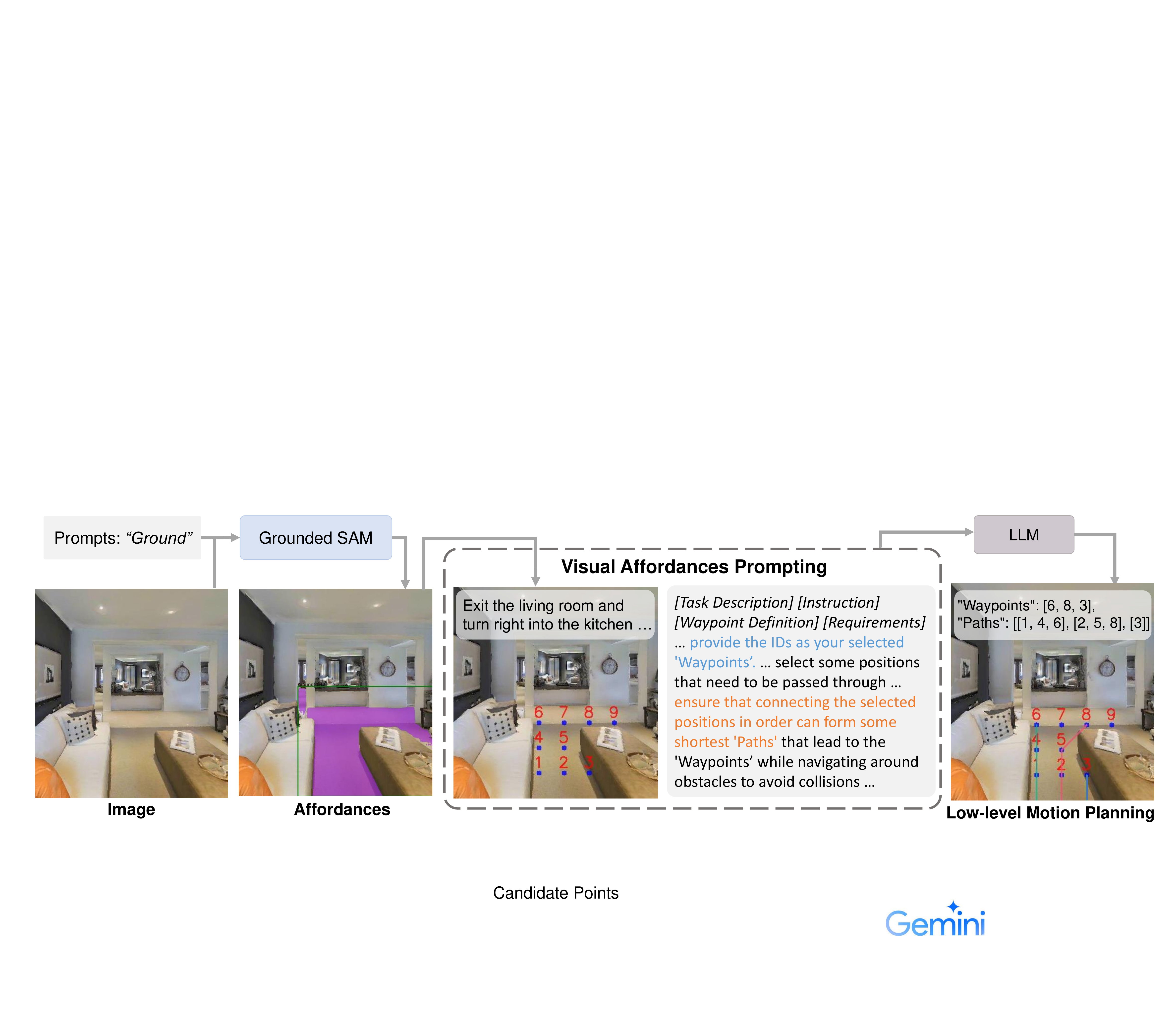}
\end{center}
\vspace{-5mm}
  \caption{
 Our proposed low-level affordances-oriented planning framework with visual affordances prompting. First, we utilize Grounded SAM to segment the visible ground as affordances. We then introduce visual affordances prompting (VAP), where we uniformly scatter  points with numeric labels within the affordances. After querying the LLM by combining the visualized new image with task definition, instruction, waypoint definition, and output requirements, we finally obtain potential waypoints and paths in this view.
  }
\vspace{-4mm}
\label{fig:low_level}
\end{figure*}

\paragraph{Navigation with Large Language Models}

In navigation tasks, there is often a wealth of visual and semantic information involved. Hence, many LLM-based navigation methods have emerged in recent years. Some methods \cite{zhou2023navgpt,long2023discuss,chen2024mapgpt} directly utilize GPT as agents and achieve zero-shot navigation through appropriate prompting methods. Another group of methods \cite{pan2023langnav,lin2024navcot,zhang2024navid,zheng2024towards}, based on open-source LLMs such as Llama \cite{touvron2023llama2}, collect data and design tasks to fine-tune the LLMs, enabling them to have impressive navigation performance while maintaining interpretability and universality. There are also methods which incorporate LLMs as part of their frameworks \cite{cai2023bridging,chen20232}, leveraging LLMs' high-level capabilities for handling subtasks such as instruction parsing, scene description, and decision-making. However, these approaches merely exploit the high-level capabilities of LLMs and still require collecting data to train low-level policies. The potentials of  LLMs in low-level motion planning remain unexplored.

\paragraph{Visual Prompting}

Prompting is an important technique for activating various capabilities of LLMs. Recently, with the development of multimodal LLMs, some visual prompting methods \cite{yang2023set,yang2023dawn,nasiriany2024pivot,lei2024scaffolding} have emerged. For example, SoM \cite{yang2023set} extracts and labels certain objects through object detection or segmentation to enhance the visual grounding capability of LLMs. \citet{yang2023dawn} systematically summarize the results and potentials achieved using different visual markers on the multimodal GPT-4V \cite{openai2023gpt4v-system,openai2023gpt4v-technical} model. These methods have demonstrated the importance of visual prompting in unleashing the multimodal understanding capabilities of LLMs. Nonetheless, they do not address motion planning problem in navigation tasks. In this paper, we propose a novel visual affordances prompting (VAP) method that allows LLMs to serve as low-level planners, making them applicable to more practical navigation tasks.

\section{Method}

\subsection{VLN-CE Task Definition}

The task of vision-language navigation can be formulated as follows. For each episode, the agent needs to follow a fixed instruction $I$ to move from the starting point to the target destination. At step $t$, the agent can obtain an observation $O_t$, which includes views $\{V_{t}^i\}_{i=1}^{N}$ from different directions. In our framework, we set $N=4$ and collect non-overlapping views from the front, back, left, and right directions as observation, i.e., $O_t = \{V_{t}^i\}_{i=1}^{4}$.
For the action space, the VLN-CE task defines four parameterized low-level actions, namely FORWARD (0.25m), ROTATE LEFT/RIGHT ($15^\circ$), and STOP. 
In this work, the agent needs to select a waypoint location and the corresponding path as an action $a_t$, and then convert the relative position into a series of low-level actions.

\subsection{Framework Overview}

\begin{figure}[t]
\begin{center}
\includegraphics[width=1.0\linewidth]{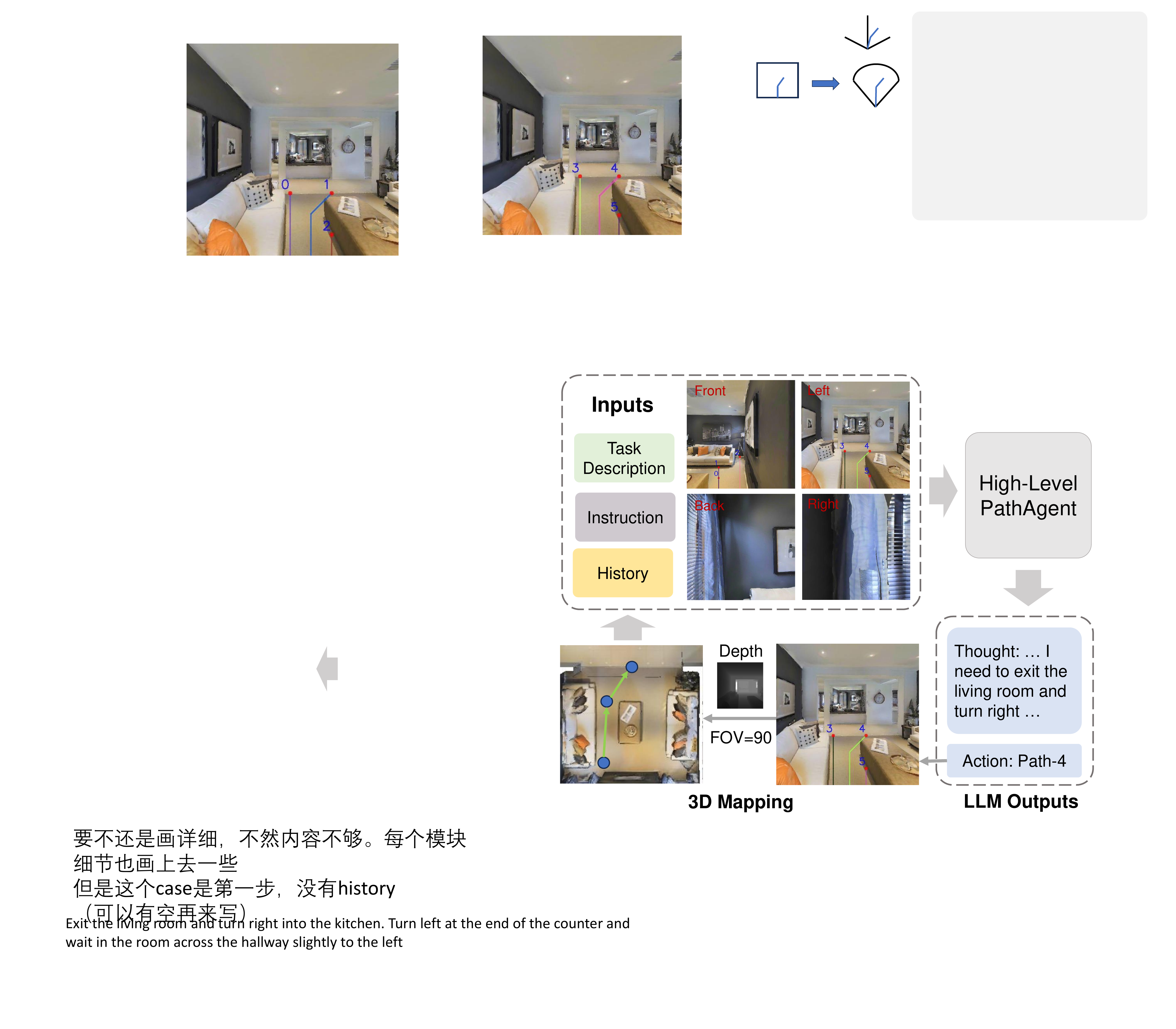}
\end{center}
\vspace{-5mm}
  \caption{
 Our proposed high-level PathAgent.
 Different from previous zero-shot VLN agents, we utilize visual prompting by marking candidate waypoints and their corresponding paths (i.e., Path 0-5) in all four observation directions. 
 This allows the PathAgent to make action decisions in the proficient RGB space and then map pixel-based paths to 3D coordinates using depth information and camera intrinsic parameters.
  }
\vspace{-3mm}
\label{fig:high_level}
\end{figure}

As shown in Figure~\ref{fig:low_level} and Figure~\ref{fig:high_level}, we follow some popular methods \cite{hong2022bridging,krantz2022sim} in the VLN-CE task and design a two-stage framework, with two LLM-based agents responsible for predicting candidate paths at the low level and making decisions based on all environmental information at the high level. This simplifies and reduces the difficulty of the task since there is no need to simultaneously address two tasks of completely different levels.

Specifically, we first utilize the Grounded SAM \cite{ren2024grounded} model to obtain navigational affordances and sample some points from them, facilitating the selection by the LLM agent. We further design prompts for the low-level agent to search for potential waypoints and plan corresponding paths by connecting the sampled points within the affordances. These candidate results are visualized and fed to the high-level agent in the second stage, where we refer to previous zero-shot methods \cite{zhou2023navgpt, chen2024mapgpt} that combine instructions and historical information to make the final decision on the location to move to. Additionally, points predicted in RGB space in the first stage can be transformed into 3D world coordinates based on depth information and camera intrinsics, and further be converted into low-level actions, thereby connecting the LLM with the 3D world.

\subsection{Visual Affordances Prompting (VAP)}
Prompting technique has been widely used to unleash the powerful capabilities of LLMs. For multimodal LLMs such as GPT-4 \cite{openai2023gpt4,openai2023gpt4v-system,openai2023gpt4v-technical} and Gemini \cite{team2023gemini,reid2024gemini}, the importance of visual prompting is gradually being recognized \cite{yang2023set,yang2023dawn,nasiriany2024pivot}. In this work, we propose a novel Visual Affordances Prompting (VAP) approach for low-level planning in continuous VLN task. As shown in Figure \ref{fig:low_level}, 
we mark a set of navigable points within affordances, from which the LLM selects candidate waypoints and connects some points to form candidate paths,
enabling low-level planning. 
VAP effectively achieves zero-shot path planning in the VLN-CE task.

\paragraph{Navigational Affordances}

``Affordances'' is a widely used concept in the field of robotics, referring to the regions in the environment where an agent's actions can be performed \cite{gibson2014ecological}. Previous works in navigation tasks \cite{qi2020learning,luddecke2017learning} have explored how to learn from supervised data and ultimately predict safe and navigable affordances in the environment. 

Fortunately, we have found that current foundation models have the capability to solve this problem in a zero-shot setting. The Grounded SAM \cite{ren2024grounded} method we adopt combines an open-set detector Grounding Dino \cite{liu2023grounding} with Segment Anything Model (SAM) \cite{kirillov2023segany}. By  inputting the prompt \textit{``Ground''}, the open-set detector can detect the visible ground and provide the corresponding bounding boxes to SAM for segmentation, resulting in  high-quality navigational affordances. 
In the environment, we set the FOV of the agent's camera to 90 degrees and collect observations from four directions in counterclockwise order, namely front, left, back, and right views.
For each view $V_t^i$ in direction $i$, the output of Grounded SAM is a set of $k$ masks $\{m_1, ... m_k\}$, and we fuse these masks into one to represent the final affordances $A_t^i$.

\paragraph{Visual Prompting}

Inspired by previous Set-of-Mark (SoM) prompting \cite{yang2023set}, we propose a novel visual affordances prompting method to assist LLM in unleashing its visual grounding and affordances-oriented spatial planning. Given a view $V_t^i$ of size $h\times w$, we uniformly distribute a set of  points within its affordances $A_t^i$. We then label these points in order and visualize them on the image, resulting in an augmented view $V_t^{i*}$. This allows us to easily query LLM to select suitable  points and provide their corresponding IDs without directly predicting their coordinates in RGB space or the 3D world.

Compared to the previous waypoint models \cite{hong2022bridging,krantz2022sim,an2024etpnav}, which only take RGB and/or depth information as inputs and directly predict  reachable locations near the current position, we incorporate semantic information and require path planning of the first-stage agent to navigate around obstacles. Specifically, we include the instructions $I$ of the current episode as part of the prompts, asking the agent to examine key information such as scene descriptions, landmarks, and objects, and to select corresponding waypoints and paths. Additionally, we provide some low-level task descriptions $D_L$, such as specific definitions for waypoints and paths,
like requiring waypoint predictions to maintain a distance from obstacles and be located in crucial regions, and connecting points to form paths that navigate around obstacles. 
The prompting process can be formulated as
\begin{equation}
    W_{t}^i, P_{t}^i = LLM(D_L, I, V_{t}^{i*}),
\end{equation}
where $i$ represents one of the four view directions. We further merge the predicted waypoints $W_t^i$ and paths $P_t^i$ from four directions into two candidate lists $W_t$ and $P_t$ for waypoints and paths at the current position.
Our proposed prompts help LLM better understand the task, enabling it to predict waypoints and plan reasonable paths based on affordances.

\subsection{High-level PathAgent}

After obtaining potential low-level waypoint and path predictions, we introduce another agent, dubbed \textbf{PathAgent}, to perform high-level decision-making. As shown in Figure~\ref{fig:high_level}, we refer to previous agents designed for discrete VLN tasks \cite{zhou2023navgpt,long2023discuss,chen2024mapgpt} and make some improvements specifically for the VLN-CE task. Instead of directly providing some captions of the visual observations or providing originally observed images to a multimodal LLM, we also utilize the idea of visual prompting. For a given step $t$, we visualize the candidate waypoints $W_t$ and paths $P_t$ predicted in the first stage by rendering them on the visual observation $O_t$ (containing four perspective directions) and marking them with IDs, resulting in an augmented observation $O_t'$. 
Through visual prompting, we only need four images to retain all observational information,
without relying on the simulator to provide images corresponding to candidate waypoints like MapGPT \cite{chen2024mapgpt}, which may neglect some observations.
Additionally, we add high-level task description $D_H$, instruction $I$, and historical information $H_t$ as inputs to the LLM. Following previous work, we require the LLM to output an interpretable thinking process $T_t$ and select one path from $P_t$ set as action $a_t$. The entire process is defined as
\begin{equation}
    T_t, a_t = LLM(D_H, I, H_t, O_t', W_t, P_t).
\end{equation}

\subsection{3D Mapping and Motion Control}
For our zero-shot framework, once the agent selects a waypoint as a subgoal, we need to convert the planned path into a sequence of actions to guide the agent along the path. 
During the low-level motion planning in the first stage, the LLM agent plans paths in the RGB space by connecting points.
A pixel-based path is actually a series of line segments. We can utilize the camera intrinsic parameters and depth information to map each point in the line segments from pixel coordinates to 3D world coordinates. For each line segment, we follow the low-level controller in ETPNav \cite{an2024etpnav}, which designs a rotate-then-forward control flow. Based on the world coordinates of two points in a line segment, we calculate their relative orientation and distance, and  convert them into a series of low-level ROTATE and FORWARD actions within the action space.
Through this approach, we bridge the gap between the LLM's predicted 2D results and the 3D world, successfully applying the spatial planning abilities of LLM to low-level motion.

\subsection{Waypoint Distillation}
In this work, we not only use LLMs with VAP as zero-shot waypoint predictors, but also explore transferring this capability to a learning-based waypoint predictor to obtain similar waypoint predictions. Specifically, we utilize our LLM-based waypoint predictor as a zero-shot data annotator to predict waypoints as pseudo ground-truths (GTs). We then leverage these GTs to train the waypoint predictor in ETPNav~\cite{an2024etpnav} and combine it with the high-level VLN agent used in ETPNav. 
By distilling the waypoint prediction capability of LLMs to a learning-based waypoint predictor, we can achieve more competitive performance as it can be combined with a learning-based high-level agent for fine-tuning, without relying on additional MP3D simulator data.
This is also a cost-effective paradigm since the scale of the pseudo labels we collect is the same as the scale of the MP3D data used by the previous waypoint predictor. We sample around 10K locations with 40K images for training, which only requires 40K calls to Gemini and a total cost of around 84 USD. However,  if directly fine-tuning the VLN agent in ETPNav using a zero-shot LLM-based predictor, we need to keep querying Gemini during this process. This probably requires 9.6M calls and costs around 20K USD.

\begin{table*}[t] 
 \centering
 \small
 \renewcommand\tabcolsep{5.0pt} 
 \begin{tabular}{llccccccccc}
 \toprule
 \multicolumn{1}{l}{\multirow{2}{*}{Settings}} & \multirow{2}{*}{Methods}    & \multirow{2}{*}{Simulator Data} & \multicolumn{4}{c}{R2R-CE} & \multicolumn{4}{c}{RxR-CE} \\ 
 \cmidrule(lr){4-7} \cmidrule(lr){8-11}
  &  &  & NE$\downarrow$ & OSR$\uparrow$ & SR$\uparrow$ & SPL$\uparrow$ & NE$\downarrow$ & SR$\uparrow$ & SPL$\uparrow$ & nDTW$\uparrow$  \\
 \midrule
 \multirow{7}{*}{Supervised} 
 & SASRA \cite{zubair2021sasra} & \checkmark  & 8.32 & - & 24 & 22 & - & - & - & - \\
 & Seq2Seq \cite{krantz2020beyond} & \checkmark  &  7.77 & 37 & 25 & 22 & 12.1 & 13.9 & 11.9 & 30.8\\
 & LAW \cite{raychaudhuri-etal-2021-language} & \checkmark & - & - & 35 & 31 & 10.9 & 8.0 & 8.0 & 38.0 \\
 & DC-VLN \cite{hong2022bridging} & \checkmark  & 6.20 & 52 & 41 & 36 & 8.98 & 27.1 & 22.7 & 46.7 \\
 & NaVid \cite{zhang2024navid} & \checkmark  & 5.47 & 49 & 37 & 36 & 8.41 & 23.8 & 21.2 & -\\
 & ETPNav \cite{an2024etpnav} & \checkmark  & 4.71 & 65 & 57 & 49 & 5.64 & 54.8 & 44.9 & 61.9 \\
 & AO-Planner (Ours) & & 5.55 & 59 & 47 & 33 & 7.06 & 43.3 & 30.5 & 50.1 \\
 \midrule
 \multirow{6}{*}{Zero-Shot} 
 & CLIP-Nav \cite{dorbala2022clip} &   & - & - & 5.6 & 2.9 & - & 9.8 & 3.2 & - \\
 & Seq CLIP-Nav \cite{dorbala2022clip} &  & - & - & 7.1 & 3.7 & - & 9.1 & 3.3 & -\\
 & Cow \cite{Gadre2022CLIPOW} & & -  & - & 7.8 & 5.8 & - & 7.9 & 6.1 & -\\
 & ZSON \cite{majumdar2022zson} & \checkmark & -  & - &  19.3 & 9.3 & - & 14.2 & 4.8 & - \\
 & $A^2$Nav \cite{chen20232} & \checkmark & -  & - & 22.6 & 11.1 & - & 16.8 & 6.3 & - \\
 & AO-Planner (Ours) & & 6.95 & 38.3 & \textbf{25.5} & \textbf{16.6} & 10.75 & \textbf{22.4} & \textbf{15.1} & 33.1 \\
 \bottomrule
 \end{tabular}
 \vspace{-2mm}
\caption{Comparison with supervised and zero-shot methods on validation unseen split of R2R-CE and RxR-CE. Our approach achieves state-of-the-art zero-shot performance using the same prompts and also obtains competitive supervised performance via distilling LLM's waypoint prediction ability into a learning-based predictor, without relying on simulator data for training low-level models. In contrast, $A^2$Nav requires sampling simulator data for training policies and ETPNav utilizes the navigation graph from discrete MP3D simulator for training a waypoint predictor.}
\vspace{-3mm}
\label{table:r2r_unseen}
\end{table*}

\section{Experiments}
\label{sec:experiments}

\subsection{Experimental Setting}

\paragraph{Dataset}
We conduct experiments on the challenging R2R-CE \cite{krantz2020beyond} and RxR-CE \cite{ku2020room} datasets. R2R-CE is derived from the discrete path annotations from the R2R dataset \cite{anderson2018vision} and is converted into continuous environments with the Habitat simulator \cite{savva2019habitat}. 
R2R-CE provides step-by-step high-level instructions which are corresponding to indoor navigation trajectories that should be followed by agents.
RxR-CE is a larger and more challenging dataset, which has longer instructions, and contains multilingual descriptions. However, its scene splits are similar to R2R-CE, and experiments are also conducted using the Habitat simulator.
We follow some previous zero-shot work \cite{zhou2023navgpt,long2023discuss,chen2024mapgpt} in the discrete VLN task, evaluating AO-Planner on the entire validation unseen set of R2R-CE and a random sampling subset with 500 cases from the validation unseen set of RxR-CE.
To save API costs, we also additionally sample a subset containing 100 cases from the validation unseen set of R2R-CE for ablation study.

\paragraph{Evaluation Metircs}
Following previous works \cite{krantz2020beyond,hong2022bridging}, we adopt some evaluation metrics widely used in the VLN tasks, i.e., navigation error (NE), oracle success rate (OSR), success rate (SR), success weighted by the inverse of the path length (SPL), and normalize dynamic time wrapping (nDTW).

\subsection{Experimental Results}

\paragraph{Zero-Shot Performance}
As shown in Table~\ref{table:r2r_unseen}, our method achieves state-of-the-art performance under zero-shot setting. It represents a 5.5\% SPL improvement on R2R-CE and an 8.8\% SPL improvement on RxR-CE over the previous best-performing method $A^2$Nav \cite{chen20232}. Furthermore, while $A^2$Nav does not require any instruction-goal pair demonstrations, it still needs to sample a large number of image-path pairs from the Habitat simulator to train low-level policies, which may suffer from potential generalization issues.
Some other methods that do not rely on simulator data show worse performance. For example, Cow \cite{Gadre2022CLIPOW} only achieves a success rate of 7.8\% and an SPL of 5.8\% on R2R-CE.
In contrast, our proposed method, while not requiring any data for training, significantly outperforms previous zero-shot methods.

Our zero-shot performance is comparable to three supervised methods, SASRA \cite{zubair2021sasra}, Seq2Seq \cite{krantz2020beyond}, and NaVid \cite{zhang2024navid} on R2R-CE or RxR-CE. However, we also observe that the current zero-shot performance still has a significant gap compared to supervised performance, which aligns with the performance comparison of zero-shot and supervised methods in the discrete VLN task \cite{zhou2023navgpt}. 
This may be due to LLMs' unfamiliarity with this VLN-CE task, lacking direct learning from the Habitat simulator. Additionally, the image quality of continuous scenes reconstructed by the Habitat simulator is not as high as images directly captured in MP3D. These factors affect the performance of LLMs.

It is important to note that we do not modify any prompts when applying AO-Planner to both R2R-CE and RxR-CE, which also demonstrates that our zero-shot method is more general and does not require additional fine-tuning like dataset-specific supervised models.

\paragraph{Distilling Waypoint Predictor}
We aim to transfer the capability of LLM-based waypoint prediction to a learning-based waypoint predictor to avoid reliance on discrete MP3D data.
For the waypoint model selection, we adopt the Chamfer distance metric used by ETPNav to evaluate the alignment between the model's waypoint predictions and the ground truths. Our waypoint predictor achieves a result of 1.08, which is slightly worse than the 1.04 achieved in the original ETPNav. This further demonstrates that using predictions from LLM as pseudo GTs can achieve high-quality waypoint predictions.

As shown in Table~\ref{table:r2r_unseen}, the experiments on R2R-CE and RxR-CE datasets demonstrate the potential of our proposed AO-Planner. By using the waypoint predictor to provide subgoals, AO-Planner significantly outperforms NaVid, which does not use any waypoint predictor, in both OSR and SR. We notice that the performance of AO-Planner is still behind ETPNav trained on MP3D ground truth. However, our method does not rely on any manual annotation or simulator data for training low-level policies.
In this work, we point out the limitations of the existing methods and re-demonstrate the effectiveness of waypoint prediction in a more practical setting. 
Our AO-Planner provides a meaningful reference by leveraging foundation model in not only serving as zero-shot predictors but also providing reliable waypoint data for training, free from the constraints of simulator data.

\begin{table}[t] 
 \centering
 \small
 \begin{tabular}{l|ccc}
 \toprule
  Prompting Methods & OSR$\uparrow$ & SR$\uparrow$ & SPL$\uparrow$  \\
  \midrule
  
  Visual Affordances Prompting & 39 & 27 & 16.9 \\
  \quad w/o Instructions & 36 & 24 & 14.6 \\
  \quad w/o Waypoint Definitions	&35&	23	&14.7 \\  
  \quad w/o Affordances &27	&15	&7.2 \\
 \bottomrule
 \end{tabular}
 \vspace{-2mm}
\caption{Ablation study on different prompting settings. 
}
 \vspace{-3mm}
\label{table:prompting}
\end{table}

\subsection{Ablation Study}

\paragraph{Different Prompting Methods}
As shown in Table~\ref{table:prompting}, we design three representative prompting methods to compare with our proposed Visual Affordances Prompting (VAP). The waypoint model designed for low-level motion in previous work \cite{hong2022bridging,krantz2022sim,an2024etpnav} typically only use RGB and/or depth information as input. Given LLM's semantic understanding capabilities, we have incorporated instructions as part of the prompts in VAP to help LLMs check scene descriptions, landmarks, and objects information. Our method with instructions obtains a 3\% improvement in SR and a 2.3\% improvement in SPL.

In VAP, we have also made some simple definitions for waypoints, including obstacle avoidance and finding crucial regions. We conduct an additional ablation without these definitions, but instead using a data style similar to MP3D, selecting waypoints that are 2-3 meters away. This results in a 4\% reduction  in SR and a 2.2\% reduction in SPL.

Another key factor in our VAP is the use of navigational affordances. 
We provide an experiment where we do not use SAM to segment the affordances, but directly scatter some points for the LLMs to select. The experiment shows that this greatly increases the difficulty for Gemini to perform low-level motion planning, as there are more options, and some points are located on obstacles, causing interference. This lowers SR from 27\% to 15\%.

\begin{table}[t] 
 \centering
 \small
\renewcommand\tabcolsep{3.0pt} 
 \begin{tabular}{l|ccc}
 \toprule
  Methods & OSR$\uparrow$ & SR$\uparrow$ & SPL$\uparrow$  \\
  \midrule
  Zero-Shot Predictor + Baseline Agent & 31 & 20 & 12.4  \\
  Supervised Predictor + Baseline Agent & 34 & 23 & 13.1 \\ 
  Zero-Shot Predictor + PathAgent & 39 & 27 & 16.9  \\
  Supervised Predictor + PathAgent & 42 & 29 & 17.6 \\
 \bottomrule
 \end{tabular}
 \vspace{-2mm}
\caption{Ablation study on different combinations of waypoint predictors and high-level VLN agents.}
 \vspace{-4mm}
\label{table:types}
\end{table}

\begin{figure*}[t]
\begin{center}
 \includegraphics[width=0.97\linewidth]{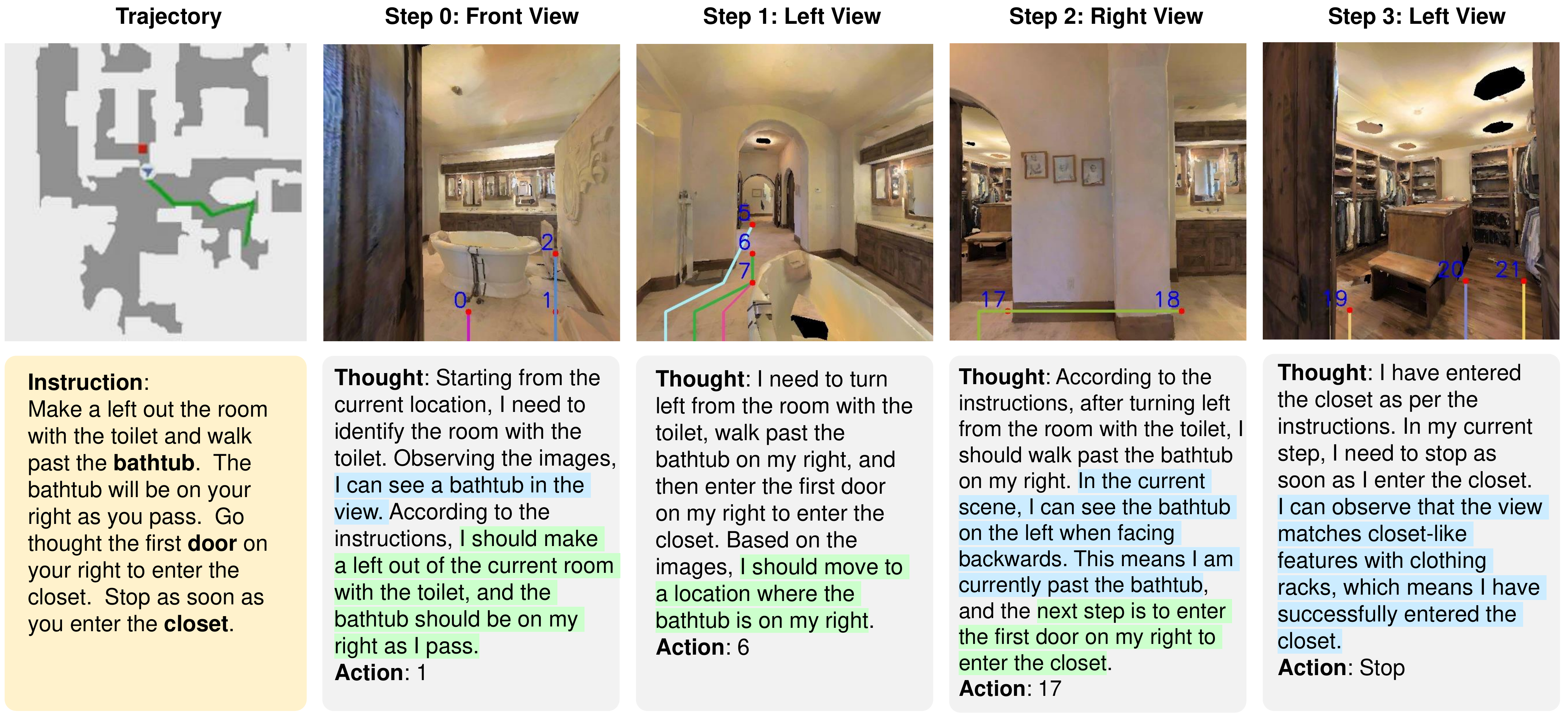}
\end{center}
\vspace{-4mm}
  \caption{
An example of successful navigation in a continuous environment. 
We present visualizations of the motion planning results from the low-level agent (upper) and the thinking process of the high-level PathAgent based on visualized candidate paths (only the selected directions are shown in this figure) and the selection of a path ID as an action (bottom). The agent ultimately decides to stop after observing the target ``closet''.
  }
\vspace{-3mm}
\label{fig:case}
\end{figure*}

\paragraph{Different Combinations of Low-Level Predictors and High-Level Agents}
Our AO-Planner defaults to using a zero-shot waypoint predictor in the first stage and our PathAgent in the second stage. In this experiment, we compare different implementations of the two stages. In Table~\ref{table:types}, ``Baseline Agent'' refers to transferring MapGPT \cite{chen2024mapgpt} from a discrete environment to the VLN-CE task. It does not visualize the waypoint predictions in the observation or calculate the position of the waypoint via 3D mapping, but simply selects a view and moves a fixed distance each time.
Our PathAgent shows a 6-7\% performance improvement in SR compared to the ``Baseline Agent''.

Besides, We explore replacing our zero-shot waypoint predictor with a supervised waypoint predictor. Our proposed zero-shot predictor does not rely on any ground-truth from the MP3D simulator for training and performs on par with (16.9\% vs. 17.6\% in SPL) the supervised waypoint predictor used by DC-VLN \cite{hong2022bridging}. This demonstrates that our visual affordances prompting is a promising new approach in zero-shot continuous VLN.

\begin{table}[t] 
 \centering
 \small
 \renewcommand\tabcolsep{4.0pt} 
 \begin{tabular}{cc|ccc}
 \toprule
 \multicolumn{2}{c|}{LLMs} & \multicolumn{3}{c}{Metircs} \\ 
  Low-Level & High-Level & OSR$\uparrow$ & SR$\uparrow$ & SPL$\uparrow$  \\
  \midrule
  GPT-4o	& Gemini-1.5-Pro	& 23	& 11	& 6.9 \\
  Gemini-1.5-Pro & Gemini-1.5-Pro & 25 & 14 & 9.1 \\
  GPT-4o & GPT-4o & 34 & 16 & 9.4 \\
  Gemini-1.5-Pro & GPT-4o & 39 & 27 & 16.9 \\
 \bottomrule
 \end{tabular}
 \vspace{-2mm}
\caption{Ablation study on using different LLMs. 
}
 \vspace{-3mm}
\label{table:llms}
\end{table}

\paragraph{Agents using Different LLMs}
Our zero-shot AO-Planner is a two-stage framework. As shown in Table \ref{table:llms}, we use different LLMs for low-level and high-level agents, respectively, to observe their impact on performance.
Gemini-1.5 has demonstrated impressive performance on some embodied tasks \cite{reid2024gemini}, such as OpenEQA \cite{majumdar2024openeqa}. In our low-level motion planning task, we find that Gemini-1.5-Pro outperforms OpenAI's GPT-4o model. However, contrary to the results of low-level ablation, Gemini-1.5 performs poorly in high-level decision-making. This could be attributed to differences in the training data of the two models, which have resulted in distinct distributions of capabilities. Investigating this underlying cause further will be our future work. Based on the aforementioned experimental results, we default to using Gemini-1.5 as the LLM for the low-level agent and GPT-4o for the high-level agent.

\subsection{Qualitative Results}

As shown in Figure \ref{fig:case}, we present a comprehensive example of navigation in a continuous environment.
It can be observed that the low-level motion planning agent provides some high-quality candidates. For instance, in the left view at step 1, the bathtub blocks most of the forward direction, but the low-level agent finds three paths that navigate around the obstacle and reach the distant waypoints. The trajectory in the top-down map also demonstrates that the agent successfully avoids the bathtub obstacle and prevents collisions.

The high-level PathAgent consistently selects the correct directions at each step, and the output thoughts also demonstrate its understanding of the task. Upon encountering landmarks that exist in the instruction, the agent promptly assesses the progress of instruction execution and makes accurate decisions until reaching the goal.

These results show the effectiveness of both our low-level motion planning and high-level PathAgent, bridging the gap between LLMs and 3D world in continuous VLN task.

\section{Conclusion}
This paper proposes AO-Planner, where we explore the previously under-researched low-level motion planning capability of LLM 
for addressing the challenging VLN-CE task, effectively bridging the gap between LLM and 3D world.
We introduce a novel visual affordances prompting strategy, which provides navigational affordances to enable LLM to plan low-level motions.
A high-level PathAgent is further proposed to perform path selection in the RGB space by comprehending all environmental information. Finally, we map the selected pixel-based path to 3D coordinates for navigation. 
Experiments on R2R-CE and RxR-CE demonstrate the effectiveness of our approach that achieves state-of-the-art zero-shot and competitive supervised performance.

\renewcommand\thesection{\Alph{section}}

\section{Implementation Details}

\subsection{Foundation Models}

We utilize multiple foundation models to build our proposed AO-Planner. The Grounded SAM \cite{ren2024grounded} we employ is an open-set segmentation model consisting of two modules: Grounded DINO \cite{liu2023grounding} generates a series of bounding boxes based on a text prompt (we use ``ground'' in our AO-Planner), while SAM \cite{kirillov2023segany} segments the objects within these bounding boxes to obtain masks. We also test multiple multimodal LLMs simultaneously for their performance in low-level and high-level planning. After conducting the ablation study, we find that the optimal configuration is Gemini-1.5-Pro \cite{reid2024gemini} for low-level motion planning, while the latest GPT-4o \footnote{https://platform.openai.com/docs/models/gpt-4o} is used for high-level decision-making.

\subsection{Prompts}
\begin{figure*}[p]
\begin{center}
 \includegraphics[width=1.0\linewidth]{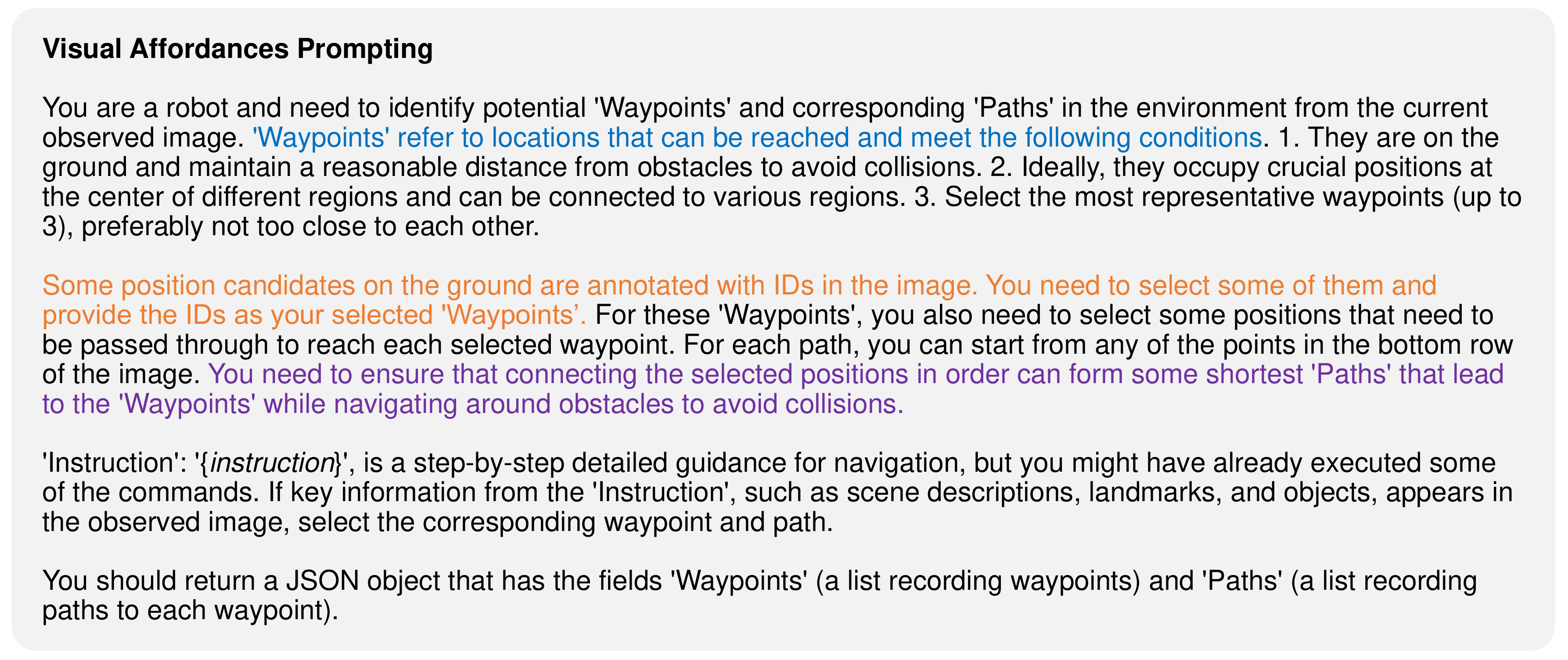}
\end{center}
  \caption{
  Task prompts for the low-level Visual Affordances Prompting (VAP).
  }
\label{fig:supp_vap}
\end{figure*}

\begin{figure*}[p]
\begin{center}
 \includegraphics[width=1.0\linewidth]{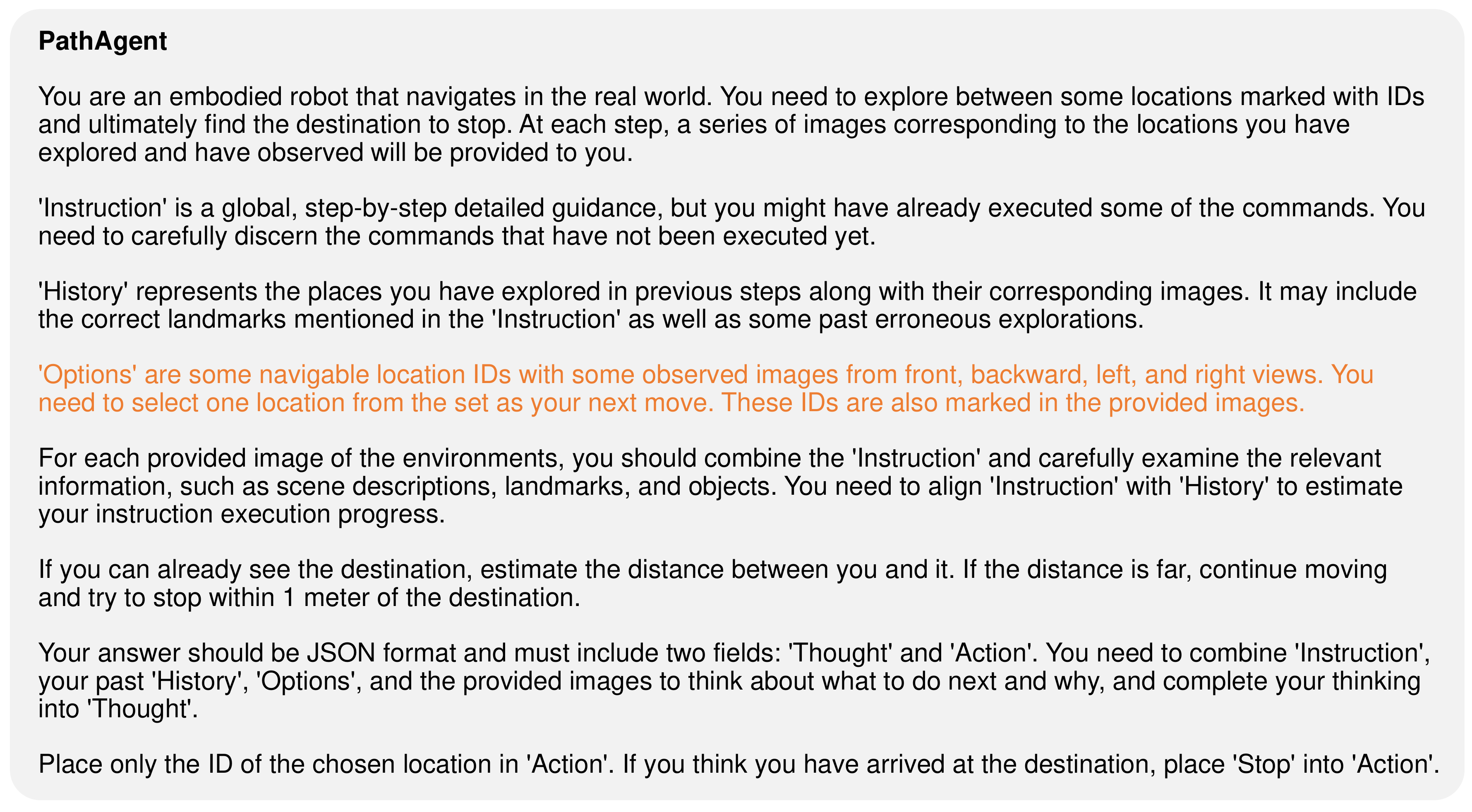}
\end{center}
  \caption{
  Task prompts for the high-level PathAgent.
  }
\label{fig:supp_path_agent}
\end{figure*}

As shown in Figure \ref{fig:supp_vap}, we list the task prompts used for Visual Affordances Prompting (VAP). In the prompts, we define the conditions that the waypoints should satisfy and explain how to find the corresponding paths while avoiding obstacles. In Figure \ref{fig:supp_path_agent}, we present task prompts for PathAgent, which are primarily based on the design by MapGPT \cite{chen2024mapgpt}, with some modifications specifically for VLN-CE. All of these are textual prompts, and our main contribution lies in visual prompting. Please refer to the main text for the design of visual prompting.

\subsection{Zero-Shot Setting}
We conduct R2R-CE and RxR-CE experiments on the Habitat \cite{savva2019habitat}. 
Our zero-shot AO-Planner is a multi-step framework. First, for the observation images in four directions, we use Grounded SAM to find the Navigational Affordances and sprinkle visual markers on some locations. These marked images, combined with Visual Affordances Prompting, can help LLMs select some points as candidate waypoints. For the second stage, the high-level agent needs to integrate various environmental and task information to select the most suitable one from these candidate waypoints. Finally, based on the camera projection principles and depth information provided by the simulator, our framework can convert the pixel coordinates to world coordinates, and then move the high-level agent to the selected position. 

\subsection{Distillation Setting}
For the distillation results, we first utilize AO-Planner to collect data and train a new waypoint predictor to replace the one used in ETPNav \cite{an2024etpnav}. We use an NVIDIA 4090 GPU to train the model and keep the hyper-parameters consistent with the publicly available predictor training script, with a batch size of 8 and a learning rate of 1e-6 using the AdamW optimizer \cite{loshchilov2018decoupled}.
We then combine this new predictor with the VLN agent adopted by ETPNav in the second stage for decision-making. Specifically, for R2R-CE, we train the model for 15,000 iterations, setting the batch size to 8, and the learning rate to 1e-5, and also use the AdamW optimizer. For RxR-CE, we change the learning rate to 1.5e-5 and adjust the number of iterations to 30,000, while keeping other hyper-parameters unchanged.

\section{More Results}

\subsection{The Accuracy of Segmentation}
To study the performance of our SAM-based open area segmentation, we configure the Habitat simulator to obtain the segmentation ground-truth for the \textit{``floor''} object. Then, we randomly sample around 1000 images and test the segmentation performance of Grounded SAM. Despite the presence of a lot of noise in the images and segmentation ground-truth obtained from Habitat, Grounded SAM still achieves a performance of 71.3\% mIoU. This result shows that our proposed approach can rely on Grounded SAM to obtain high-quality affordances for navigation tasks.

\subsection{Reasons for Failure}
We manually check 100 failed cases and analyze the reasons for failure. Around 70\% of the failures are due to mistakes in the decision-making by the high-level agent, and the remaining 30\% are due to issues in low-level planning. This also shows the effectiveness of our core design, i.e., the low-level planner. Concretely, the failures of high-level agents primarily lie in choosing the wrong direction, with some others stopping prematurely at the incorrect location or missing the target and continuing to explore. The failures of low-level planning mainly lie in detecting walls and tables as part of the ground, leading to incorrect affordances segmentation, or selecting waypoints that are too far, resulting in some collisions. 

\subsection{The Alignment between Thoughts and Actions}
We manually check the thoughts and actual actions taken by the high-level PathAgent in 100 random steps. There are only 2 instances where the thoughts do not mention the specific action information, and only 1 instance where the action mentioned in the thoughts is inconsistent with the final action taken. This shows that the alignment between agent's thoughts and actions is generally well in our AO-Planner.

\bibliography{aaai25}

\end{document}